\documentclass[letterpaper, 10 pt, conference]{ieeeconf}  

\IEEEoverridecommandlockouts                              

\usepackage{graphics} 
\usepackage{epsfig} 
\usepackage{mathptmx} 
\usepackage{times} 
\usepackage{amsmath} 
\usepackage{amssymb}  
\usepackage{siunitx}
\usepackage{booktabs}
\usepackage{censor}
\usepackage{hyperref}
\usepackage{makecell}
\usepackage{float}
\usepackage{dblfloatfix}

\usepackage{caption}
\newcommand{\improve}[1]{\underline{#1}}
\newcommand{\bestimprove}[1]{\underline{\mathbf{#1}}}
\title{\LARGE \bf
RAFL: Generalizable Sim-to-Real of Soft Robots with \\ Residual Acceleration Field Learning
\\[0.5em]
\large \href{https://generalroboticslab.com/RAFL}{generalroboticslab.com/RAFL}
}

\author{Dong Heon Cho$^{1}$ and Boyuan Chen$^{1,2,3}$
\thanks{*This work was supported by ARO under award W911NF2410405, and by DARPA TIAMAT program under award HR00112490419.}
\thanks{$^{1}$Department of Computer Science, Duke University, 308 Research Dr., Durham, 27705, NC, USA}
\thanks{$^{2}$Department of Mechanical Engineering and Material Science, 121 Hudson Hall, Durham, 27708, NC, USA}
\thanks{$^{3}$Department of Electrical and Computer Engineering, 100 Science Dr., Durham, 27708, NC, USA}
}

\begin{document}

\maketitle
\begingroup
\renewcommand\thefootnote{}
\footnotetext{Code is available at: \href{https://github.com/generalroboticslab/RAFL.git}{https://github.com/generalroboticslab/RAFL.git}}
\endgroup
\thispagestyle{empty}
\pagestyle{empty}

\begin{abstract}

Differentiable simulators enable gradient-based optimization of soft robots over material parameters, control, and morphology, but accurately modeling real systems remains challenging due to the sim-to-real gap. This issue becomes more pronounced when geometry is itself a design variable. System identification reduces discrepancies by fitting global material parameters to data; however, when constitutive models are misspecified or observations are sparse, identified parameters often absorb geometry-dependent effects rather than reflect intrinsic material behavior. More expressive constitutive models can improve accuracy but substantially increase computational cost, limiting practicality.

We propose a residual acceleration field learning (RAFL) framework that augments a base simulator with a transferable, element-level corrective dynamics field. Operating on shared local features, the model is agnostic to global mesh topology and discretization. Trained end-to-end through a differentiable simulator using sparse marker observations, the learned residual generalizes across shapes. In both sim-to-sim and sim-to-real experiments, our method achieves consistent zero-shot improvements on unseen morphologies, while system identification frequently exhibits negative transfer. The framework also supports continual refinement, enabling simulation accuracy to accumulate during morphology optimization.

\end{abstract}


\section{INTRODUCTION}

Differentiable simulators have become a central tool for inverse problems in soft robotics, enabling gradient-based optimization over control, material parameters, and increasingly, morphology itself \cite{review_diffSim}. When geometry is treated as a design variable, simulation allows rapid iteration over candidate shapes without fabrication, while computed gradients accelerate convergence toward optimized designs.



In many applications, differentiable simulation is first calibrated to match the behavior of a physical system, for example through system identification of material parameters \cite{diffSimID, sim2real-softRobot}. Such calibration can yield accurate models for the observed geometry and operating conditions. However, a key challenge arises when the morphology itself changes. Parameters identified on one geometry do not necessarily generalize to new designs, and simulation accuracy may degrade when applied to unseen shapes or discretizations. When constitutive models are misspecified or observations are sparse, the identification problem becomes structurally underdetermined. Fitted parameters absorb geometry-dependent effects, yielding effective quantities that are conditioned on a particular shape or discretization rather than reflecting intrinsic material properties. As a result, parameters identified on one morphology often fail to transfer reliably to new geometries, requiring repeated tuning during morphology exploration.

Residual physics learning \cite{resPhys} offers an alternative by augmenting a base simulator with learned corrective dynamics. Yet existing residual formulations typically operate on global state representations. Because the learned correction is tied to a specific topology such as fixed degrees of freedom, transfer across geometries and discretizations still remains limited. In practice, both system identification and global residual models entangle correction with mesh structure, limiting their applicability for robust sim-to-real transfer that can generalize to various morphologies.

\begin{figure}[!t]
      \centering
      \setlength{\fboxrule}{0pt}
      \framebox{\parbox{3in}{
      \includegraphics[scale=1.0]{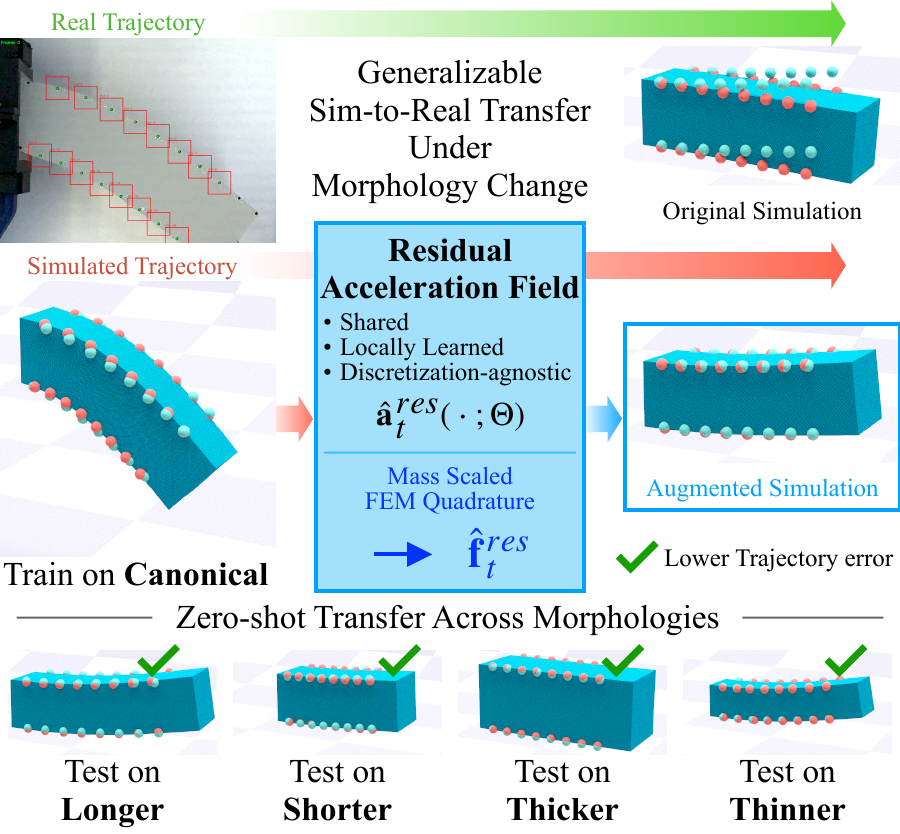} }}
      \caption{RAFL overview. Our generalizable residual acceleration field framework is trained end-to-end to reduce error on recorded sparse marker trajectories, and is agnostic to mesh topology, allowing zero-shot transfer to unseen morphologies.}
      \label{teaser}
\end{figure}

We propose a residual acceleration field learning (RAFL) framework (Fig.~\ref{teaser}) that shifts the level of learning from the global state to shared element-level features that are locally defined and agnostic to discretization. 
By predicting local corrective accelerations from physically structured deformation and rate features, and accumulating them through standard finite element quadrature, the model is not bound to a particular mesh or global state dimension, enabling cross-geometry transfer.

We evaluated this framework in both sim-to-sim and sim-to-real settings involving cantilever beams and fish-tail geometries with varying dimensions, topology, and mesh resolution. A model trained on a single canonical beam demonstrated consistent zero-shot improvements on unseen morphologies, while traditional system identification frequently exhibited negative transfer. Moreover, our residual acceleration field can be continually refined as new morphologies are observed, allowing simulation accuracy to accumulate rather than reset across designs. This capability supports rapid evaluation of new candidate geometries without sacrificing prior sim-to-real calibration, which provides a scalable pathway toward morphology-robust differentiable simulation.

\section{RELATED WORK}

Recent advances in differentiable simulation have enabled efficient gradient-based optimization of complex systems, achieving faster convergence compared to gradient-free methods \cite{review_diffSim, review_diffSoftRobot}. In soft-body robotics, differentiable simulators \cite{diffpd, difftaichi, diffxpbd} have supported applications in manipulation \cite{fluidlab, dexDeform, diffcloth}, control \cite{starfish, chainqueen}, and morphology optimization \cite{diffaqua, softzoo}, where geometry itself becomes a decision variable.

Despite these advances, the sim-to-real gap remains a central obstacle. Many morphology optimization works \cite{chainqueen, diffaqua, softzoo, aquaROM} still rely on fixed simulator dynamics and do not explicitly address cross-geometry sim-to-real consistency. As a result, while performance may be optimistic within simulation, physically realized geometries may fail to replicate the same behavior. To our knowledge, prior work has not explicitly demonstrated consistent cross-geometry sim-to-real transfer under both topology and discretization changes.

A common strategy for reducing the sim-to-real gap is system identification \cite{real2sim, gradsim, sim2realSoftFish}, which fits global material parameters to observed trajectories. However, under sparse observations or model misspecification, identification becomes underdetermined due to ``parameter observability'' \cite{diffSimID}, as multiple parameter settings can explain the same data. In practice, fitted parameters absorb \mbox{geometry-,} \mbox{discretization-,} and environment-dependent effects \cite{sim2real-softRobot}, yielding morphology-conditioned effective quantities that fail to transfer reliably across shapes. While iterative tuning approaches \cite{tunenet} can help adapt parameters to new settings, they require additional data for each configuration, which limits its efficiency in morphology exploration.

Another line of work learns constitutive laws directly from data using collections of prior models or large language models \cite{omniphysgs, sga}. Although these methods can improve simulation fidelity for specific materials, they remain limited when sim-to-real discrepancies arise from behaviors not represented within the assumed constitutive family, such as internal damping or unmodeled dissipation. Moreover, increasing constitutive expressiveness often amplifies stiffness and computational cost, reducing practicality within large-scale differentiable optimization loops.

Neural augmentation of simulation pipelines provides an alternative strategy. Learning corrective dynamics through additive residual forces \cite{resPhys} or flow matching \cite{conditionalFlow} has been shown to improve simulation accuracy while preserving computational efficiency. However, such residual formulations operate on fixed global state representations or specific mesh topologies, restricting transfer across geometries and discretizations.

Physics-informed neural architectures further improve data efficiency and structural inductive bias by replacing specific components of the simulator with learned modules. For example, past work~\cite{fastAquaticSwimmer} has modeled external hydrodynamic forces via a physics-constrained neural network \cite{fluidModel} to capture complex fluid–structure interactions efficiently. Others \cite{nclaw, uniphy} have replaced internal constitutive laws with lightweight neural models that are not explicitly tied to a single morphology. Nevertheless, these approaches typically target particular components of the dynamics rather than providing a general residual correction mechanism capable of absorbing arbitrary unmodeled effects while remaining topology-agnostic.

Rather than replacing the full system dynamics, we aim to learn transferable residual corrective dynamics that preserve the structure of the base simulator. Related ideas have appeared in hybrid PDE solvers and neural operator literature, where learned corrections augment coarse or misspecified numerical models to improve solution fidelity \cite{hybridPDE}. Unlike approaches that seek to approximate full PDE solution operators \cite{neuralOperator-acc, neuralOperator}, we focus on transferable sim-to-real correction that augments existing solutions and generalizes across varying morphologies and mesh resolutions. Additionally, our framework operates solely on local features, without the need for full system encoding. Consequently, our framework further relaxes the dependence on training morphology.

\section{METHOD}

\begin{figure*}[t]
  \centering
  \setlength{\fboxrule}{0pt}
  \framebox{\parbox{6in}{
  \includegraphics[scale=1.0]{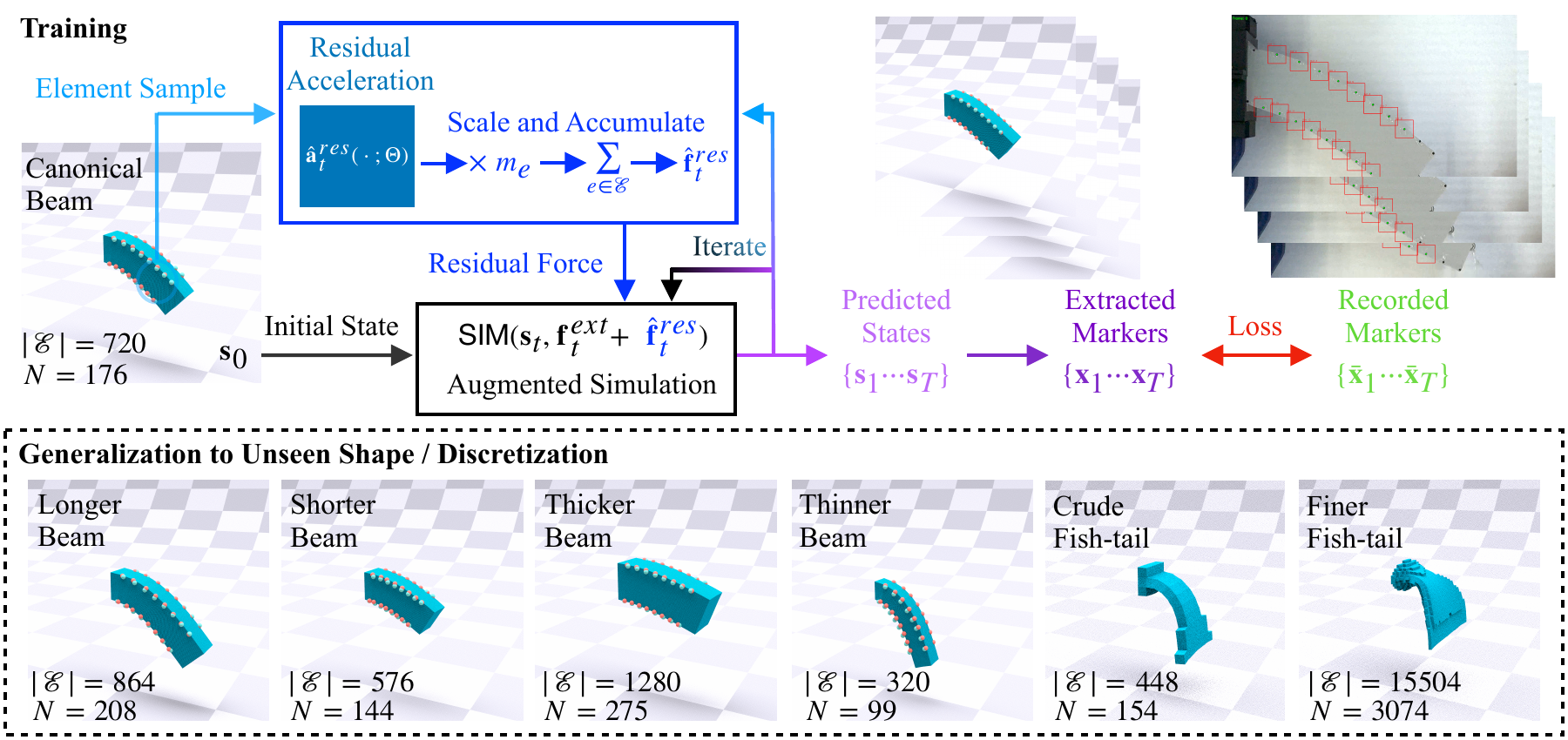} }}
  \caption{RAFL pipeline overview. The network predicts element-level residual accelerations from local deformation/rate features, which are mass-scaled and accumulated via FEM quadrature to form residual forces. Trained end-to-end from sparse marker trajectories, the model transfers zero-shot across morphologies and discretizations.}
  \label{pipeline}
\end{figure*}

Following prior residual force learning work \cite{resPhys}, we augment a base simulator with corrective forces applied at each time step. However, directly learning residual nodal forces ties the correction to training mesh topology. To remove this structural dependence, we instead learn a quadrature-defined residual acceleration \emph{field} evaluated locally in material space. The full pipeline is visualized in Fig.~\ref{pipeline}.

\subsection{Simulation Preliminaries}

We simulate soft bodies using a corotational linear elastic material model in DiffPD~\cite{diffpd}. While simplified, this choice provides a stable and efficiently differentiable base model, which is subsequently augmented by our learned residual dynamics to capture complex real-world behavior. 

Each body is discretized as a hexahedral mesh with $N$ vertices, such that at time step $t$, the state $\mathbf{s}_t$ consists of vertex positions and velocities, $\mathbf{q}_t, \mathbf{v}_t \in \mathbb{R}^{N \times 3}$. From the full state, an operator $\mathbf{x}(\mathbf{s}_{t})$ extracts observable markers corresponding to recorded data $\bar{\mathbf{x}}_{t}$.

Given external forces $\mathbf{f}^{ext}_t \in \mathbb{R}^{N \times 3}$, the simulator advances the system with time step $\Delta t$ as $\mathbf{s}_{t+1} = \textsc{Sim}(\mathbf{s}_t, \mathbf{f}^{ext}_t)$ by solving the following system of equations:
\begin{equation}\label{1}
\begin{split}
\mathbf{q}_{t+1} &= \mathbf{q}_t + \Delta t \mathbf{v}_{t+1}, \\
\mathbf{v}_{t+1} &= \mathbf{v}_t + \Delta t \mathbf{M}^{-1}
\left[ \mathbf{f}^{int}(\mathbf{q}_{t+1}) + \mathbf{f}^{ext}_t \right],
\end{split}
\end{equation}
where $\mathbf{M}$ is the mass matrix and $\mathbf{f}^{int}$ denotes internal elastic forces defined by the material model. In our passive setting, the $\mathbf{f}^{ext}_t = \mathbf{M}\mathbf{g}$ corresponds to gravity. 

A residual model 
$\hat{\mathbf{f}}^{res}(\boldsymbol{\phi}_t; \Theta) \in \mathbb{R}^{N \times 3}$ 
parameterized by weights $\Theta$ augments the simulation such that
\begin{equation}\label{eq_augmented_sim}
\mathbf{s}_{t+1}
= \textsc{Sim}\left(\mathbf{s}_t, \mathbf{f}^{ext}_t + \mathbf{\hat{f}}^{res}(\boldsymbol{\phi}_t \, ; \Theta) \right).
\end{equation}
where the input $\boldsymbol{\phi}_t = (\mathbf{s}_t, [\mathbf{f}^{ext}_t])$ optionally includes  $\mathbf{f}^{ext}_t$.

Internal quantities are evaluated over 8-point Gaussian quadrature element samples $\mathcal{E}$. Each sample $e \in \mathcal{E}$ has mass $m_e$, volume $V_e$, shape functions $\mathbf{N}_e \in \mathbb{R}^{E}$ and shape function gradients $\nabla_X \mathbf{N}_e \in \mathbb{R}^{E \times 3}$. Given the $E = 8$ surrounding nodal positions $\mathbf{x}_e \in \mathbb{R}^{E \times 3}$ 
and velocities $\mathbf{v}_e \in \mathbb{R}^{E \times 3}$, the
deformation gradient and velocity gradient are computed as
\begin{equation}\label{2}
\begin{split}
\mathbf{F}_e &= \sum_{i=1}^{E} \mathbf{x}_{e,i} \otimes \nabla_X \mathbf{N}_{e,i}, \\
\mathbf{L}_e &= \left( \sum_{i=1}^{E} \mathbf{v}_{e,i} \otimes \nabla_X \mathbf{N}_{e,i} \right) \mathbf{F}_e^{-1}.
\end{split}
\end{equation}
From $\mathbf{L}_e$, we obtain the strain rate tensor $\mathbf{D}_e = \tfrac{1}{2}(\mathbf{L}_e + \mathbf{L}_e^T)$ and the spin tensor $\mathbf{W}_e = \tfrac{1}{2}(\mathbf{L}_e - \mathbf{L}_e^T)$. Element-wise external body forces are given by $m_e \mathbf{g}$.

Following prior residual learning work~\cite{resPhys}, our passive beams were fabricated from Smooth-On Dragon Skin 10 (Shore 10A). Similarly, the base simulator used density $\rho = 1070\,\si{kg\,m^{-3}}$, Young’s modulus $E = 215\,\si{kPa}$, and Poisson’s ratio $\nu = 0.45$ for numerical stability. 

\subsection{Supervised Residual Baseline}

A natural formulation for learning residual physics is to model residual nodal forces directly as a function of the global state by using a single large neural network $\mathbf{\hat{f}}^{res}(\mathbf{s}_t;\Theta)$\cite{resPhys}. However, this approach exhibits two fundamental limitations.

First, possibly due to the high-dimensional optimization landscape, we found that a global-state model cannot be trained end-to-end through the differentiable simulator on recorded trajectories. Instead, residual targets are pre-collected via per-step optimization:
\begin{equation}\label{eq_collected_fres}
{\mathbf{f}_t^{res}}^* =
\arg \min_{\mathbf{f}_t^{res}}
\Big(
\|\mathbf{x}(\mathbf{s}_{t+1}) - \bar{\mathbf{x}}_{t+1}\|^2_2
+ \lambda_1 \|\mathbf{f}_t^{res}\|^2_2
\Big).
\end{equation}
where the second term regularizes correction magnitude.
After collecting all target residual forces ${\mathbf{f}_t^{res}}^*$ and corresponding input states $\mathbf{s}_t$, the network is trained separately to minimize 
\begin{equation}\label{eq_supervised}
\mathcal{L} :=
\sum_{i \in \text{batch}}
\| \mathbf{\hat{f}}^{res}(\mathbf{s}_i;\Theta) - {\mathbf{f}_i^{res}}^*\|^2_2
+ \lambda_2 \|\Theta\|^2_2.
\end{equation}
with L2 weight decay on the network parameters. This expensive two-step process limits efficient training across multiple shapes.

More importantly, because forces are defined over discrete vertices, they remain topology-dependent, and thus the model can not generalize to different discretizations. To evaluate this supervised residual learning (SRL) baseline from Gao et al. \cite{resPhys} across geometries without architectural modification, we maintained a constant hexahedral grid and varied shape via anisotropic scaling of internal elements. 

\subsection{Residual Acceleration Field Learning}

\begin{table*}[t]
\renewcommand{\arraystretch}{0.88}
\setlength{\tabcolsep}{2pt}
\setlength{\aboverulesep}{0pt}
\setlength{\belowrulesep}{0pt}
\setlength{\abovetopsep}{0pt}
\setlength{\belowbottomsep}{0pt}
\caption{Input Features Provided to the Network}
\label{tab1}
\begin{center}
\begin{tabular}{ccc}
\toprule
\textbf{Feature} & \textbf{Dimension} & \textbf{Description}\\
\midrule
$\textsc{diag}(\boldsymbol{\Sigma}_e) - 1$ & 3 & Principal Stretches centered by 1 \\
$\textsc{flattened}(\mathbf{F}_e^T\mathbf{F}_e - I)$ & 9 & Right Cauchy Green Tensor centered by identity matrix \\
$\det(\mathbf{F}_e) - 1$ & 1 & Determinant of $\mathbf{F}_e$ centered by 1\\
$\textsc{diag}(\mathbf{R}_e^T\mathbf{D}_e\mathbf{R}_e)$ & 3 & Strain Rates along principal axes (co-rotated to rest frame)\\
$\textsc{trace}(\mathbf{R}_e^T\mathbf{D}_e\mathbf{R}_e)$ & 1 & Volumetric Strain Rate \\
$\textsc{flattened}(\mathbf{R}_e^T\mathbf{D}_e\mathbf{R}_e - \frac{1}{3}\textsc{trace}(\mathbf{R}_e^T\mathbf{D}_e\mathbf{R}_e)I)$ & 9 & Shear Strain Rates (co-rotated to rest frame)\\
$\sqrt{\frac{1}{2} \|\mathbf{R}_e^T\mathbf{D}_e\mathbf{R}_e - \frac{1}{3}\textsc{trace}(\mathbf{R}_e^T\mathbf{D}_e\mathbf{R}_e)I\|^2_F}$ & 1 & Magnitude of Shear Strain Rate \\
$\textsc{axial}(\mathbf{R}_e^T\mathbf{W}_e\mathbf{R}_e)$ & 3 & Spin Rates (co-rotated to rest frame)\\
$\|2*\textsc{axial}(\mathbf{R}_e^T\mathbf{W}_e\mathbf{R}_e)\|_2$ & 1 & Vorticity Magnitude \\
$\mathbf{R}_e^Ta^{ext}_e$ & 3 & Acceleration due to external forces (i.e. gravity $a^{ext}_e = \mathbf{g}$) (co-rotated to rest frame)\\
\bottomrule
\end{tabular}
\end{center}
\end{table*}

To remove discretization dependence, we reformulate residual learning as the prediction of a continuous acceleration field $\hat{\mathbf{a}}^{res}(\mathbf{d}_{t,e}\,; \Theta) \in \mathbb{R}^3$ evaluated at each quadrature sample $e \in \mathcal{E}$. 
The element-level features $\mathbf{d}_{t,e} \in \mathbb{R}^{34}$ are computed from the deformation and velocity gradients in \eqref{2} and external accelerations, as summarized in Tab.~\ref{tab1}.  

Predicted accelerations are then scaled by element mass and accumulated via quadrature to obtain the residual force:
\begin{equation}\label{eq_accumulated_fres}
\hat{\mathbf{f}}^{res}(\mathbf{s}_t, \mathbf{f}^{ext}_t; \Theta)
= \sum_{e \in \mathcal{E}} 
\mathbf{N}_e^T \, m_e \, \mathbf{R}_e \, 
\hat{\mathbf{a}}^{res}(\mathbf{d}_{t,e}; \Theta).
\end{equation}
This design decouples the correction from mesh structure, enabling transfer across discretizations without architectural modification.

Inspired by prior work in neural constitutive laws \cite{nclaw}, our network architecture enforces two structural constraints. First, rotational equivariance is achieved by extracting the rotation 
$\mathbf{R}_e$ from the polar decomposition 
$\mathbf{F}_e = \mathbf{U}_e \boldsymbol{\Sigma}_e \mathbf{V}_e^T$, with $\mathbf{R}_e = \mathbf{U}_e \mathbf{V}_e^T$. Tensor-valued features are co-rotated into the local rest frame prior to inference, and the predicted residual acceleration is rotated back 
to the spatial frame using $\mathbf{R}_e$. Second, zero residual output at rest is enforced by removing bias terms and centering all input features so that they vanish in the stationary undeformed configuration under zero external loading.


Notably, our model can be trained end-to-end through the differentiable simulator by minimizing the trajectory error obtained by iterating the augmented simulation defined in \eqref{eq_augmented_sim}. The parameters $\Theta$ are optimized over $R$ trajectories of length $T$:
\begin{equation}
\begin{split}
\Theta^* = \arg \min_{\Theta}
\frac{1}{R(T-1)}
\sum_{r=1}^R \sum_{t=0}^{T-2}
\Big(
&\|\mathbf{x}(\mathbf{s}^r_{t+1}) - \bar{\mathbf{x}}^r_{t+1}\|^2_2\\
&+ \lambda_1 \|\mathbf{\hat{f}}^{res}(\mathbf{s}_t, \mathbf{f}^{ext}_t; \Theta)\|^2_2
\Big) \\
&+ \lambda_2 \|\Theta\|^2_2.
\end{split}
\end{equation}
where the initial state $\mathbf{s}_0^r$ is provided to initialize each trajectory $r$.  
This direct training framework, with the same correction-magnitude and weight-decay regularizers as SRL, further enables efficient fine-tuning on different meshes without requiring expensive residual target preprocessing.

In all our experiments, we utilized a multi-layer perceptron (MLP) architecture with 4 layers and 64 hidden units per layer, containing significantly fewer parameters \mbox{($\approx 0.02 \,\%$)} compared to the SRL baseline. The network was trained using Adam for 100 epochs with early stopping based on validation trajectories. 

\subsection{Sim-to-Sim Setting}

We first evaluated in a sim-to-sim setting using a target simulator $\textsc{Sim}_{target}$ with altered material parameters. In this setting, full vertex states were available. As a result, the operator $\mathbf{x}(\mathbf{s}_t)$ and the corresponding ground truth observations $\bar{\mathbf{x}}_t$ were simply given by the simulated vertex positions $\mathbf{q_t}$ from the augmented simulation and from $\textsc{Sim}_{target}$, respectively.

We evaluated all trained models on a collection of $S$ held-out test trajectories of length $T$, using the error metric
\begin{equation}\label{sim2sim_evaluation}
\mathcal{E}_q =
\frac{1}{S T N}
\sum_{r,t,i}
\| \mathbf{q}^r_{t,i} - \bar{\mathbf{q}}^r_{t,i} \|_2 .
\end{equation}

Using the same error metric in \eqref{sim2sim_evaluation}, we evaluated generalization under two regimes: 1) anisotropic scaling of a fixed-topology mesh, and 2) genuinely different shapes and discretizations with varying element counts. 
The latter regime was only applicable to our element-level formulation.

\subsection{Sim-to-Real Setting}

In the \textit{sim-to-real} setting, full-state information was not available. Instead, we utilized a sparse set of physical markers distributed across the surface of the object. In this case, the observation operator $\mathbf{x}(\mathbf{s}_t)$ corresponded to the interpolated positions of these markers obtained from nearby surface vertices using barycentric coordinates, as proposed in prior work \cite{resPhys}. 


Performance was measured by
\begin{equation}\label{sim2real_evaluation}
\mathcal{E}_x =
\frac{1}{S T M}
\sum_{r,t,i}
\| \mathbf{x}^r_{t,i} - \bar{\mathbf{x}}^r_{t,i} \|_2 ,
\end{equation}
where $M$ denotes the number of markers, and $\mathbf{x}^r_{i,t}$ and $\bar{\mathbf{x}}^r_{i,t}$ correspond to the simulated and recorded marker positions, respectively.

Initial cross-shape evaluation mirrored the sim-to-sim protocol. To further assess generalization in the sim-to-real setting, we adopted an expanded evaluation protocol. For each morphology, we trained on trajectories from that configuration and evaluated \eqref{sim2real_evaluation} on all others. 
We omitted the SRL baseline in this expanded setting due to the substantial dataset generation cost required for each morphology. 
In contrast, our framework does not require pre-dataset collection to provide supervised residual targets, making it significantly more scalable across physical configurations.

\subsection{System Identification Baseline}

As another baseline, we performed system identification (SysID) by optimizing Young’s modulus $E$ and Poisson’s ratio $\nu$ to minimize trajectory reconstruction error:
\begin{equation}
\mathcal{L}_s =
\frac{1}{R(T-1)}
\sum_{r,t}
\|\mathbf{x}(\textsc{Sim}(\mathbf{s}^r_t, \mathbf{f}^{ext}_t; E,\nu)) - \bar{\mathbf{x}}^r_{t+1}\|^2_2 .
\end{equation}

In addition to jointly optimizing $(E,\nu)$, we also considered a simplified variant in which only the Young's modulus $E$ was optimized while fixing the Poisson's ratio to $\nu = 0.499$ to reflect the nearly incompressible nature of the material. 

Cross-shape generalization followed the same expanded protocol used for RAFL. Our expanded cross-shape generalization experiments reported in the main results reflect the full SysID setting where both $E$ and $\nu$ were optimized. We observed that the identified Young's modulus varied across shapes when $\nu$ was fixed. 
This behavior suggests that parameters fitted to one morphology may absorb geometry-dependent effects rather than reflect intrinsic material properties, leading to reduced transferability across shapes.

\section{RESULTS}

We evaluated clamped soft beams across multiple geometries, including a canonical beam with dimension \(10\times3\times3\,\si{cm}\) used for training, as well as longer/shorter \((12\times3\times3\,\si{cm}/\ 8\times3\times3\,\si{cm})\) and thicker/thinner \((10\times4\times4\,\si{cm}/\ 10\times2\times2\,\si{cm})\) beams for testing generalization. In the sim-to-sim experiments, we additionally evaluated two (crude/ finer) fish-tail meshes to assess transfer beyond beam-like morphologies.

Canonical beam simulations employed $1\,\si{cm}^3$ hexahedral elements. Scaled geometries were generated by anisotropic scaling while preserving mesh topology. Non-scaled geometries together with the crude fish-tail, used the same element size but varied the element count. The finer fish-tail used \(3\,\si{mm}^3\) elements to evaluate generalization under substantial discretization changes. Fig.~\ref{pipeline} visualizes all meshes, along with element sample and vertex counts of the non-scaled cases; scaled cases retain the canonical sample count (720) and number of vertices (176). 


\subsection{Sim-to-Sim Transfer}

We first evaluated sim-to-sim transfer by matching a target simulator under an identical discretization. Tab.~\ref{table_sim2sim_experiments} summarizes the material parameters used in the original and target simulations. 

\begin{table}[!ht]
\caption{\mbox{Sim-to-Sim Experimental Configuration}}
\label{table_sim2sim_experiments}
\begin{center}
\setlength{\tabcolsep}{2pt}
\setlength{\aboverulesep}{0pt}
\setlength{\belowrulesep}{0pt}
\setlength{\abovetopsep}{0pt}
\setlength{\belowbottomsep}{0pt}
\renewcommand{\arraystretch}{0.95}
\begin{tabular}{ccc}
\toprule
\textbf{Simulation} & \textbf{Young's Modulus} & \textbf{Poisson's Ratio} \\
\midrule
Original & $215 \,\si{kPa}$ & $0.45$\\
Target & $264 \,\si{kPa}$ & $0.499$\\
\bottomrule
\end{tabular}
\end{center}
\end{table}

Two passive motion regimes were considered. The first involved vibrating beams generated by releasing the beam from equilibrium under an attached tip weight sampled within $[50, 220]\,\si{g}$. The second involved twisted beams produced by releasing the beam from initial twist angles sampled within $[\pi/6, \pi]$. 
Tab.~\ref{table_sim2sim_data} details the datasets used in these experiments. 

\begin{table}[!ht]
\caption{Sim-to-Sim Datasets}
\label{table_sim2sim_data}
\begin{center}
\setlength{\tabcolsep}{2pt}
\setlength{\aboverulesep}{0pt}
\setlength{\belowrulesep}{0pt}
\setlength{\abovetopsep}{0pt}
\setlength{\belowbottomsep}{0pt}
\renewcommand{\arraystretch}{0.95}
\begin{tabular}{ccc}
\toprule
& \textbf{Vibrating Beam} & \textbf{Twisted Beam}  \\
\midrule
Step size $(\si{s})$ & 0.01 & 0.01\\
Time steps & 150 & 100 \\
Training Trajectories  & 9 & 10 \\
Validation Trajectories  & 2 & 2\\
Testing Trajectories  & 5 & 8 \\
\bottomrule
\end{tabular}
\end{center}
\end{table}

The number of training and validation trajectories only pertain to the canonical beam and the two fish-tail shapes, where fine-tuning was performed. The number of testing trajectories indicates the data used to evaluate each geometry.

Both the SRL baseline and RAFL models were trained exclusively on canonical beam trajectories and applied zero-shot to all other geometries and discretizations. Tab.~\ref{table_vibrating_sim2sim} and \ref{table_twist_sim2sim} report simulation errors as in \eqref{sim2sim_evaluation}. 
Entries where residual models improve upon the original simulator are underlined, and the lowest error in each row is bolded.

\begin{table}[!ht]
\caption{Vibrating Beam Sim-to-Sim Results}
\label{table_vibrating_sim2sim}
\begin{center}
\setlength{\tabcolsep}{2pt}
\setlength{\aboverulesep}{0pt}
\setlength{\belowrulesep}{0pt}
\setlength{\abovetopsep}{0pt}
\setlength{\belowbottomsep}{0pt}
\renewcommand{\arraystretch}{0.95}
\begin{tabular}{cccc}
\toprule
& \multicolumn{3}{c}{\textbf{Simulation Error ($\si{mm}$)}} \\
\textbf{Geometry} & \textbf{Original} & \textbf{RAFL (ours)} & \textbf{SRL (Gao et al.)} \\
\midrule
Canonical & $4.369  \pm 0.761 $ & $\improve{0.058  \pm 0.045}$ & $\bestimprove{0.023  \pm 0.083}$\\
Longer (Scaled) & $8.518  \pm 1.681 $& $\bestimprove{1.016  \pm 0.344}$ & $\improve{1.078  \pm 0.329}$\\
Shorter (Scaled) & $1.907  \pm 0.277 $& $\bestimprove{0.288 \pm 0.058}$ & $\improve{0.433  \pm 0.079}$\\
Thicker (Scaled) & $2.656  \pm 0.347 $& $\bestimprove{0.302  \pm 0.061}$ & $\improve{1.022  \pm 0.162}$\\
Thinner (Scaled) & $8.658  \pm 2.916 $& $\bestimprove{1.358  \pm 0.591}$ & $17.644  \pm 11.550 $\\
Longer  & $8.163 \pm 1.760 $& $\bestimprove{1.441  \pm 0.528}$ & N/A\\
Shorter  & $2.018  \pm 0.267 $& $\bestimprove{0.401  \pm 0.077}$ & N/A\\
Thicker  & $2.345  \pm 0.297 $& $\bestimprove{0.156  \pm 0.036}$ & N/A\\
Thinner  & $10.212  \pm 3.124$& $\bestimprove{1.043  \pm 0.435}$ & N/A\\
Crude Fish-tail & $16.078  \pm 4.624 $& $\bestimprove{7.836  \pm 3.577}$ & N/A\\
Finer Fish-tail & $7.777  \pm 2.615 $& $\bestimprove{3.101  \pm 1.860}$ & N/A\\
\bottomrule
\end{tabular}
\end{center}
\end{table}
\begin{table}[!ht]
\caption{Twisted Beam Sim-to-Sim Results}
\label{table_twist_sim2sim}
\begin{center}
\setlength{\tabcolsep}{2pt}
\setlength{\aboverulesep}{0pt}
\setlength{\belowrulesep}{0pt}
\setlength{\abovetopsep}{0pt}
\setlength{\belowbottomsep}{0pt}
\begin{tabular}{cccc}
\toprule
& \multicolumn{3}{c}{\textbf{Simulation Error ($\si{mm}$)}} \\
\textbf{Geometry} & \textbf{Original} & \textbf{RAFL (ours)} & \textbf{SRL (Gao et al.)} \\
\midrule
Canonical & $4.360  \pm 1.551 $ & $\improve{0.130  \pm 0.192}$ & $\bestimprove{0.022 \pm 0.044}$\\
Longer (Scaled) & $8.384  \pm 4.130 $& $\bestimprove{0.525  \pm 0.445}$ & $\improve{0.963  \pm 0.710}$\\
Shorter (Scaled) & $1.915  \pm 0.427 $& $\improve{0.148 \pm 0.177}$ & $\bestimprove{0.108  \pm 0.041}$\\
Thicker (Scaled) & $2.659  \pm 0.761 $& $\bestimprove{0.239  \pm 0.184}$ & $\improve{0.973  \pm 0.295}$\\
Thinner (Scaled) & $8.389  \pm 4.071 $& $\bestimprove{0.575  \pm 0.522}$ & $\improve{3.245  \pm 1.180}$\\
Longer  & $8.033 \pm 4.095 $& $\bestimprove{0.799  \pm 0.522}$ & N/A\\
Shorter  & $2.021  \pm 0.461 $& $\bestimprove{0.193  \pm 0.140}$ & N/A\\
Thicker  & $2.354  \pm 0.618 $& $\bestimprove{0.491  \pm 0.180}$ & N/A\\
Thinner  & $9.960  \pm 5.184$& $\bestimprove{1.998  \pm 1.058}$ & N/A\\
Crude Fish-tail & $16.471  \pm 10.610 $& $\bestimprove{6.070  \pm 6.073}$ & N/A\\
Finer Fish-tail & $7.634  \pm 3.201 $& $\bestimprove{4.531  \pm 4.962}$ & N/A\\
\bottomrule
\end{tabular}
\end{center}
\end{table}

On the canonical beam, both methods substantially reduced error relative to the original simulator. SRL achieved the lowest error on the training geometry, while RAFL remained comparable despite the enforced locality constraints.

Across scaled geometries, RAFL achieved the lowest error in most cases and remained stable even when the baseline exhibited strong degradations. For example, on the thinner scaled vibrating beam, SRL increased the error to $17.644\,\si{mm}$ while RAFL reduced it to $1.358\,\si{mm}$. For non-scaled geometries, where the SRL baseline is inapplicable due to its fixed global state dimensionality, RAFL consistently reduced error across all discretizations. Although absolute errors were higher for the fish-tail geometries than for the beam-like cases, zero-shot transfer still yielded improvements over the original simulator, indicating that our learned acceleration field captures local correction structure that can transfer beyond the training morphology.

To evaluate adaptability, we fine-tuned the RAFL models on fish-tail trajectories, which yielded substantial improvements as shown in Tab.~\ref{table_finetune_sim2sim}. 
These results suggest that the residual acceleration field provides a strong transferable initialization requiring only modest adaptation to specialize to new morphologies.

\begin{table}[!ht]
\caption{Sim-to-Sim Fish-tail Fine-tuning Results}
\label{table_finetune_sim2sim}
\centering
\setlength{\tabcolsep}{2pt}
\setlength{\aboverulesep}{0pt}
\setlength{\belowrulesep}{0pt}
\setlength{\abovetopsep}{0pt}
\setlength{\belowbottomsep}{0pt}
\renewcommand{\arraystretch}{0.95}
\begin{tabular}{ccccc}
\toprule
 & \multicolumn{4}{c}{\textbf{Simulation Error ($\si{mm}$)}}\\
 & \multicolumn{2}{c}{\textbf{Vibrating Beam}} & \multicolumn{2}{c}{\textbf{Twisted Beam}}\\
\textbf{Geometry} & \textbf{Zero-shot} & \textbf{Fine-tuned} & \textbf{Zero-shot} & \textbf{Fine-tuned}\\
\midrule
Crude Fish-tail
& \makecell{$\improve{7.836}$ \\ $\improve{\pm 3.577}$}
& \makecell{$\bestimprove{0.784}$ \\ $\bestimprove{\pm 0.122}$}
& \makecell{$\improve{6.070}$ \\ $\improve{\pm 6.073}$}
& \makecell{$\bestimprove{2.252}$ \\ $\bestimprove{\pm 4.405}$}\\

Finer Fish-tail
& \makecell{$\improve{3.101}$ \\ $\improve{\pm 1.860}$}
& \makecell{$\bestimprove{0.467}$ \\ $\bestimprove{\pm 0.148}$}
& \makecell{$\improve{4.531}$ \\ $\improve{\pm 4.962}$}
& \makecell{$\bestimprove{1.464}$ \\ $\bestimprove{\pm 2.667}$}\\
\bottomrule
\end{tabular}
\end{table}

\subsection{Sim-to-Real Transfer}

We evaluated sim-to-real transfer using sparse surface markers collected from physical cantilever beam experiments. Consistent with the vibrating sim-to-sim setting, we considered a clamped cantilever beam, where various initial weights between $[50\,\si{g},220\,\si{g}]$ were applied to the free end, and released from the corresponding equilibrium state. Videos were recorded at 100\,fps and surface markers were tracked and reconstructed in 3D using a calibrated orthographic model. Tab.~\ref{table_sim2real_data} outlines the datasets used in these experiments. 
For each trajectory, the initial simulator configuration was estimated by optimizing a sequence of virtual forces following the procedure described in prior work~\cite{resPhys}.

\begin{table}[!ht]
\caption{Sim-to-Real Datasets}
\label{table_sim2real_data}
\begin{center}
\setlength{\tabcolsep}{2pt}
\setlength{\aboverulesep}{0pt}
\setlength{\belowrulesep}{0pt}
\setlength{\abovetopsep}{0pt}
\setlength{\belowbottomsep}{0pt}
\renewcommand{\arraystretch}{0.95}
\begin{tabular}{cc}
\toprule
Number of Markers & 32\\
Step size $(\si{s})$ & 0.01 \\
Time steps & 140 \\
Training Trajectories  & 9  \\
Validation Trajectories  & 2 \\
Testing Trajectories  & 5  \\
\bottomrule
\end{tabular}
\end{center}
\end{table}

Tab.~\ref{table_sysid_parameters} reports the material parameters identified on the canonical beam. 
The estimates remained in a physically plausible range and improved accuracy on the training geometry, confirming that SysID is a competitive baseline.

\begin{table}[!ht]
\caption{Sim-to-Real Material Parameters}
\label{table_sysid_parameters}
\begin{center}
\setlength{\tabcolsep}{2pt}
\setlength{\aboverulesep}{0pt}
\setlength{\belowrulesep}{0pt}
\setlength{\abovetopsep}{0pt}
\setlength{\belowbottomsep}{0pt}
\renewcommand{\arraystretch}{0.95}
\begin{tabular}{ccc}
\toprule
\textbf{Training Type} & \textbf{Young's Modulus} & \textbf{Poisson's Ratio} \\
\midrule
None (Original) & $215 \,\si{kPa}$ & $0.45$\\
SysID (both) & $234 \,\si{kPa}$ & $0.476$\\
SysID (Young's Modulus only) & $116 \,\si{kPa}$ & $0.499$ (Fixed)\\
\bottomrule
\end{tabular}
\end{center}
\end{table}

Tab.~\ref{table_vibrating_sim2real} compares performance across all geometries using the error metric in \eqref{sim2real_evaluation}. Improvements over the original simulator are underlined and the lowest error per row is bolded. 

\begin{table*}[t]
\caption{Vibrating Beam Sim-to-Real Results}
\label{table_vibrating_sim2real}
\setlength{\aboverulesep}{0pt}
\setlength{\belowrulesep}{0pt}
\setlength{\abovetopsep}{0pt}
\setlength{\belowbottomsep}{0pt}
\begin{center}
\begin{tabular}{cccccc}
\toprule
& & &\textbf{Simulation Error $(\si{mm})$}& &\\
\textbf{Geometry} & \textbf{Original} & \textbf{SysID (both)} & \textbf{SysID (Young's Modulus only)} & \textbf{RAFL (ours)} & \textbf{SRL (Gao et al.)} \\
\midrule
Canonical & $1.903  \pm 2.021 $ 
& $\improve{1.704  \pm 2.035}$ 
& $\improve{1.894  \pm 1.981}$ 
& $\improve{0.646  \pm 0.467}$ 
& $\mathbf{0.523  \pm 0.318}$\\
Longer (Scaled) & $3.627  \pm 3.138 $ 
& $\improve{3.250  \pm 3.206}$ 
& $\improve{3.544  \pm 3.082}$ 
& $\bestimprove{3.196  \pm 1.754}$ 
& $3.987  \pm2.731 $\\
Shorter (Scaled) & $1.643  \pm 1.100 $ 
& $2.027 \pm 1.051$ 
& $1.978  \pm 1.051$ 
& $\bestimprove{1.042  \pm 0.723}$ 
& $\improve{1.173  \pm 0.803}$\\
Thicker (Scaled) & $1.050  \pm 0.804 $ 
& $1.553  \pm 0.795$ 
& $1.470  \pm 0.780$ 
& $\improve{0.836  \pm 0.405}$ 
& $\bestimprove{0.835  \pm 0.678}$\\
Thinner (Scaled) & $5.807  \pm 5.593 $ 
& $\improve{5.525  \pm 5.442}$ 
& $\improve{5.736  \pm 5.195}$ 
& $\bestimprove{5.604  \pm 3.414}$ 
& $7.533  \pm 5.184 $\\
Longer  & $3.703 \pm 3.108 $ 
& $\bestimprove{3.277  \pm 3.179}$ 
& $\improve{3.552  \pm 3.030}$ 
& $\improve{3.499  \pm 1.938}$ 
&  N/A\\
Shorter  & $1.625  \pm 1.109 $ 
& $2.028  \pm 1.057$ 
& $2.110  \pm 1.038$ 
& $\bestimprove{1.053  \pm 0.759}$ 
&  N/A\\
Thicker  & $1.337  \pm 0.807 $ 
& $1.825  \pm 0.779$ 
& $1.725  \pm 0.776$ 
& $\bestimprove{0.657  \pm 0.231}$ 
&  N/A\\
Thinner  & $6.571  \pm 5.415$ 
& $\improve{5.779  \pm 5.327}$ 
& $\improve{5.837  \pm 5.097}$ 
& $\bestimprove{3.536  \pm 1.774}$ 
&  N/A\\
\bottomrule
\end{tabular}
\end{center}
\end{table*}

On the canonical beam, all methods improved upon the original simulator. The SRL baseline achieved the lowest error on the training geometry, while RAFL remained competitive and substantially outperformed both SysID variants.

A clear distinction emerged in cross-shape transfer. Both SysID methods and the SRL baseline exhibited unstable performance with negative transfer, occasionally improving performance but frequently degrading performance when applied to unseen geometries. For the SRL baseline, this instability appeared in the scaled setting, and SysID additionally degraded in the non-scaled setting. Joint optimization of $(E,\nu)$ improved robustness slightly compared to fitting $E$ alone, but neither variant consistently generalized across shapes.

In contrast, RAFL improved upon the original simulator in every scaled and non-scaled case, and achieved the lowest error in most settings. For example, on the non-scaled thinner beam, the error decreased from $6.571\,\si{mm}$ to $3.536\,\si{mm}$, substantially outperforming both SysID baselines.

\begin{figure}[!ht]
      \centering
      \setlength{\fboxrule}{0pt}
      \framebox{\parbox{3in}{
      \includegraphics[scale=1.0]{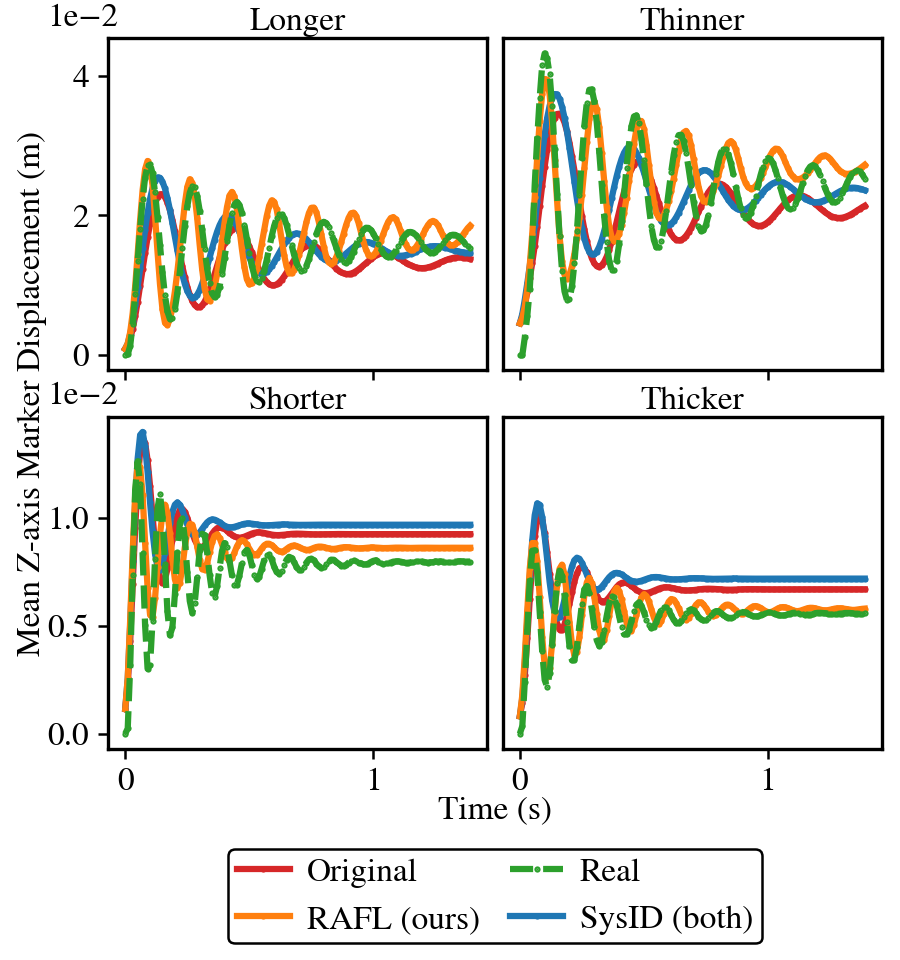} }}
      \caption{Example generalization trajectories for non-scaled geometries. SysID suffered from negative transfer, where the tuned simulation (i.e. blue curve) showed worse performance compared to the original simulation (i.e. red curve). Our framework showed consistent improvement in all generalization cases, and showed more qualitatively accurate oscillatory behavior within the trajectories.}
      \label{generalization_trajectories}
\end{figure}

Fig.~\ref{generalization_trajectories} shows representative trajectories for the non-scaled cases. While joint SysID maintained reasonable accuracy for longer and thinner beams, it degraded noticeably for shorter and thicker beams. RAFL yielded consistent improvements and more accurate long-horizon behavior across all cases.

\begin{figure*}[p]
  \centering
  \setlength{\fboxrule}{0pt}
  \includegraphics[width=\textwidth,height=0.95\textheight,keepaspectratio]{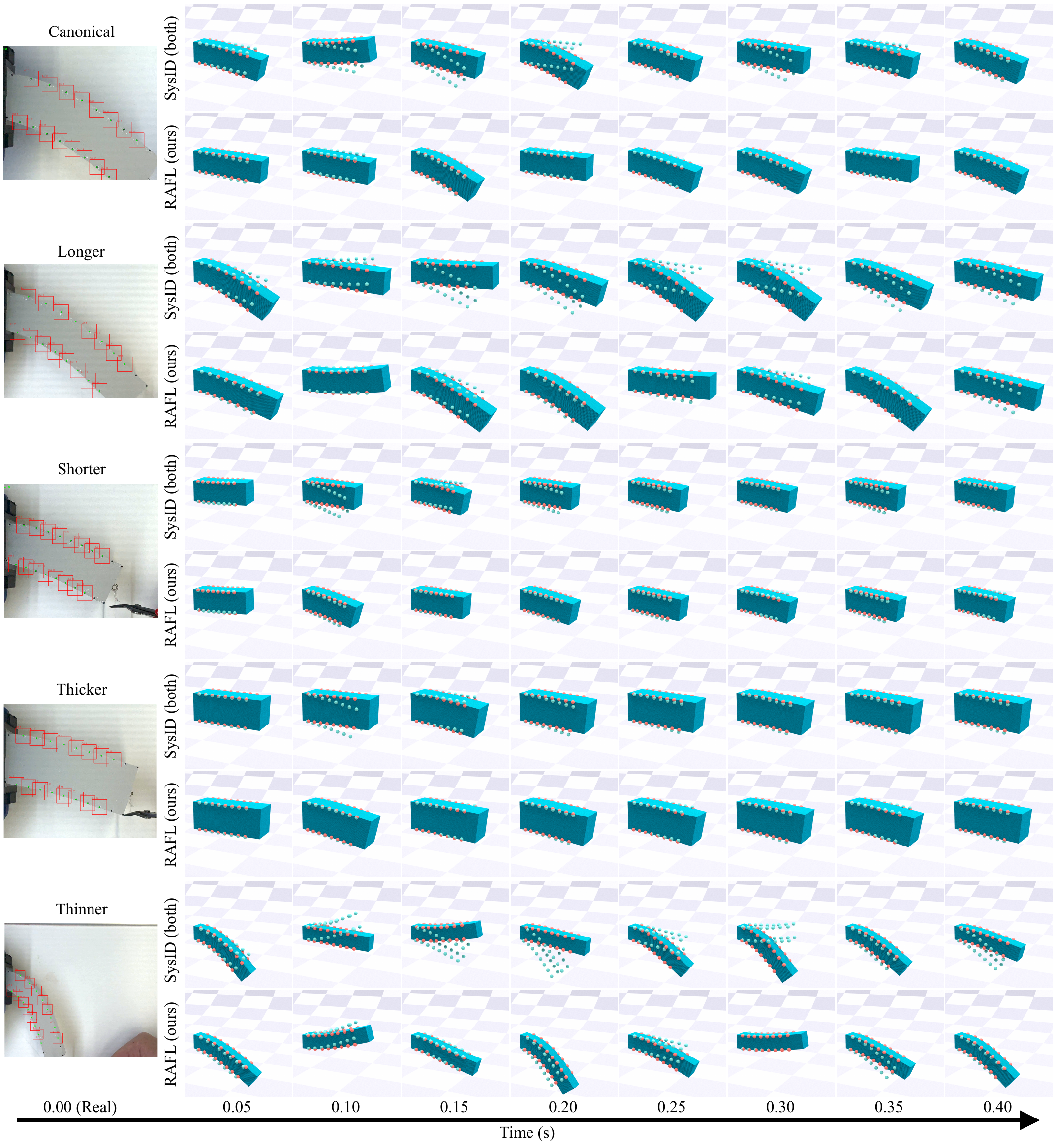} 
  \caption{Sim-to-real trajectory visualization across beam geometries. The leftmost frame in each row shows the initial state from the recorded real-world video, with surface markers highlighted by a red bounding box. Subsequent panels show simulated frames over the first 0.4 seconds (sampled every 5 frames). Rows compare SysID (both) and our RAFL model. Blue spheres indicate measured marker positions from the real system, while red spheres indicate simulated marker positions.
}
  \label{trajectories_visualized}
\end{figure*}

Fig.~\ref{trajectories_visualized} presents frames from the first $0.4$ seconds of the representative sim-to-real test trajectories for the non-scaled geometries, sampled every 5 frames. Compared to SysID, RAFL more accurately reproduces the oscillatory motion observed in the physical system.

\begin{figure}[!ht]
  \centering
  \setlength{\fboxrule}{0pt}
  \framebox{\parbox{3in}{
  \includegraphics[scale=1.0]{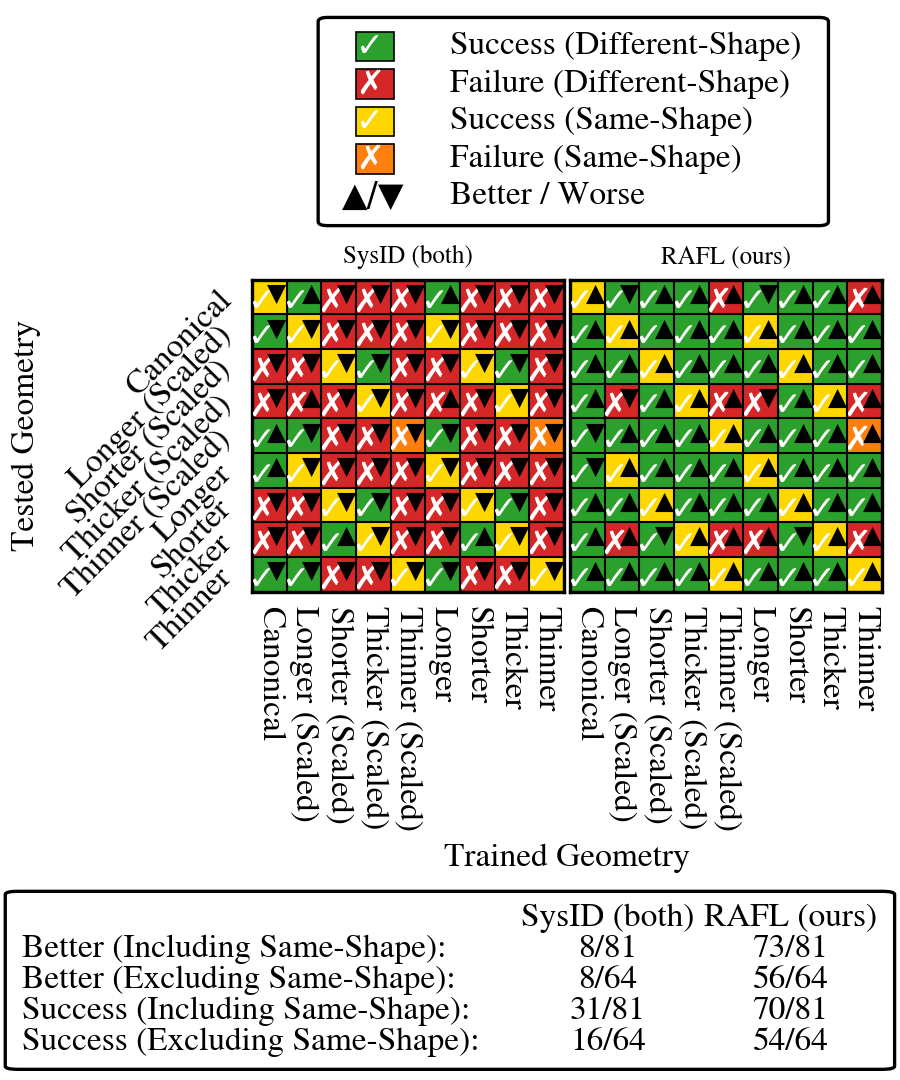} }}
  \caption{Expanded cross-shape generalization results. SysID frequently failed to transfer between morphologies, leading to negative transfer. In contrast, RAFL successfully demonstrated zero-shot accuracy improvement between meshes in most cases. RAFL also achieved better accuracy than SysID in most settings.}
  \label{expanded_generalization}
\end{figure}

The expanded evaluation in Fig.~\ref{expanded_generalization}
further highlights this trend. 
SysID frequently exhibited negative transfer (i.e. Failure) across morphology pairs, whereas RAFL consistently achieved zero-shot improvements (i.e. Success). Overall, these results demonstrate that learning transferable local residual accelerations enables more reliable sim-to-real generalization than globally optimized material parameters.

\subsection{Residual Force Structure and Transfer Behavior}

\begin{figure}[!ht]
  \centering
  \setlength{\fboxrule}{0pt}
  \framebox{\parbox{3in}{
  \includegraphics[scale=1.0]{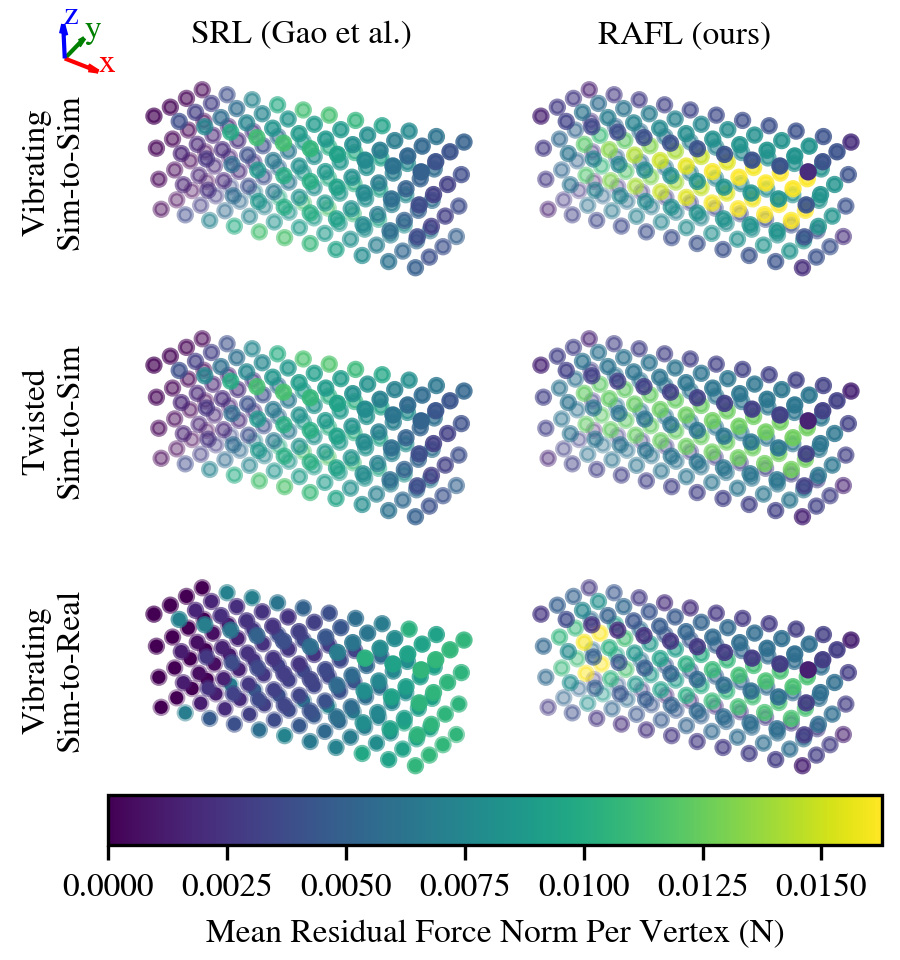} }}
  \caption{Mean residual force norm per vertex. Residual forces collected for SRL are highly dependent on vertex position compared to those learned implicitly through RAFL.}
  \label{residual_force_norms}
\end{figure}

To better understand the observed differences in transfer behavior, we analyzed the spatial distributions of residual forces produced from \eqref{eq_collected_fres} by the SRL baseline and those derived implicitly by RAFL. Fig.~\ref{residual_force_norms} visualizes the mean residual force norm per vertex across all trajectories of the canonical beam for the three experimental settings.

The SRL baseline exhibited spatially localized residuals, primarily concentrated near surfaces and the free end~(i.e. +x). This pattern indicates a strong dependence on vertex position and suggests that the learned residuals are tightly coupled to the specific mesh geometry. In contrast, the residual forces induced by our method were more uniformly distributed throughout the beam interior. This behavior suggests reduced sensitivity to vertex location and weaker coupling to global geometric structure. Such differences in residual force structure are consistent with the improved zero-shot transfer performance observed across geometries and discretizations.

\section{CONCLUSION AND FUTURE WORK}

We presented a generalizable residual acceleration field framework for improving soft-body simulation accuracy across different morphologies. By predicting local, element-level residual accelerations from physically structured features and accumulating them through standard finite element quadrature, our approach decouples corrective dynamics from mesh structure. This design mitigates the geometry–material entanglement commonly observed in system identification and avoids the discretization dependence of global residual force models. Across sim-to-sim and sim-to-real experiments, a model trained on a single beam geometry transferred reliably to unseen geometries and discretizations, yielding consistent zero-shot improvements. In contrast, system identification frequently exhibited negative transfer under morphology changes. We further demonstrated that the learned residual field provides a strong transferable initialization that can be efficiently refined on new morphologies in a fully end-to-end differentiable manner, enabling simulation accuracy to accumulate rather than reset across designs.

These findings suggest a scalable route for closing the sim-to-real gap in morphology optimization. Candidate geometries can be evaluated with a reusable, discretization-agnostic correction model that generalizes across shapes and can be incrementally updated as new data becomes available.

Several directions remain for future work. Extending the framework to actuated soft robots and contact-rich scenarios will require incorporating control inputs and more complex interaction forces into the residual formulation. While the current architecture enforces 
rotational structure and zero residual behavior at rest, additional physical priors, such as stability, passivity, or energy-consistency constraints, may further improve robustness. Finally, embedding our framework to co-evolve morphology optimization and sim-to-real calibration can offer a promising direction for scalable soft robot design.

\section*{ACKNOWLEDGMENT}

The authors thank Yuxuan Sun for assistance with fabrication of the soft beam specimens used in the experiments.

\bibliographystyle{IEEEtran}
\bibliography{references}

\newpage
\section*{APPENDIX}

\setcounter{figure}{0}
\renewcommand{\thefigure}{A\arabic{figure}}

This appendix provides additional quantitative error analysis and qualitative visualizations of the generalization behavior discussed in the main text.

\begin{figure}[!ht]
  \centering
  \setlength{\fboxrule}{0pt}
  \framebox{\parbox{3in}{
  \includegraphics[scale=1.0]{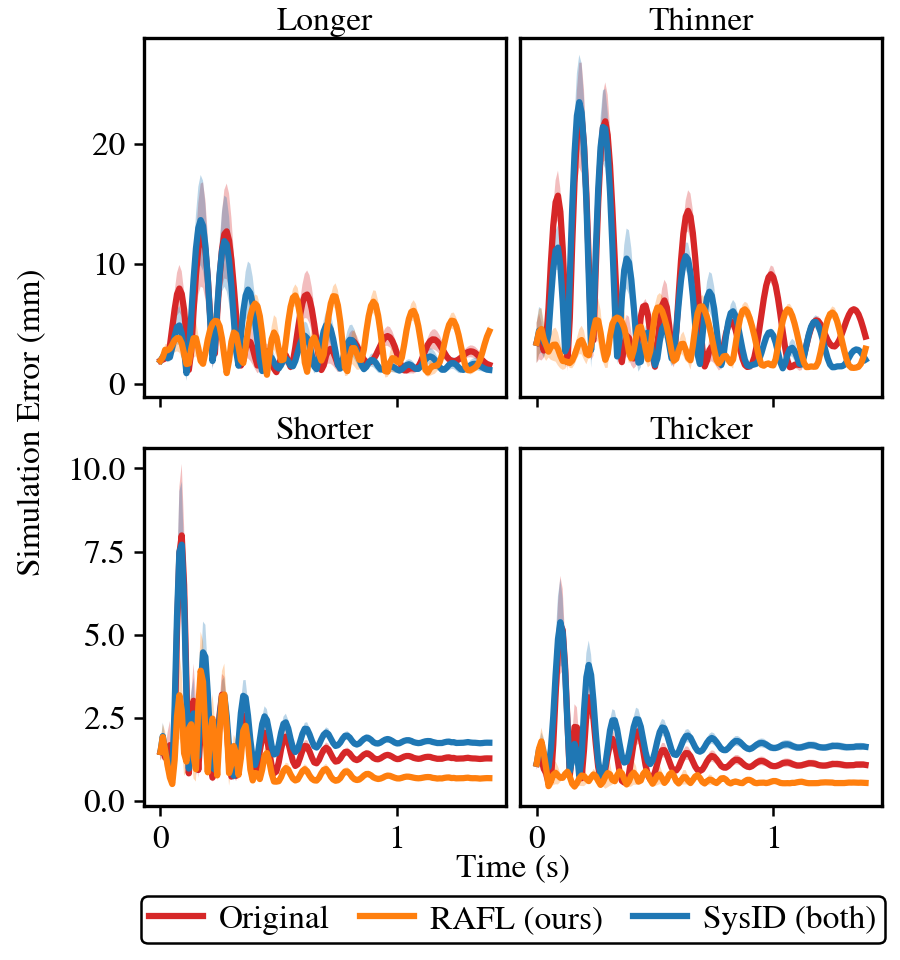} }}
  \caption{Mean marker errors of the generalization trajectories for non-scaled geometries. SysID suffers from negative transfer with greater mean marker error than the untuned simulation in the shorter and thicker beams.}
  \label{generalization_errors}
\end{figure}

Fig.~\ref{generalization_errors} compares the mean marker errors across the original simulation, SysID, and RAFL in the sim-to-real generalization experiments. While SysID improves performance on the training geometry, it frequently exhibits negative transfer when applied to unseen geometries. In contrast, RAFL consistently achieves lower marker error across all generalization cases.

\begin{figure}[H]
  \centering
  \setlength{\fboxrule}{0pt}
  \framebox{\parbox{3in}{
  \includegraphics[scale=1.0]{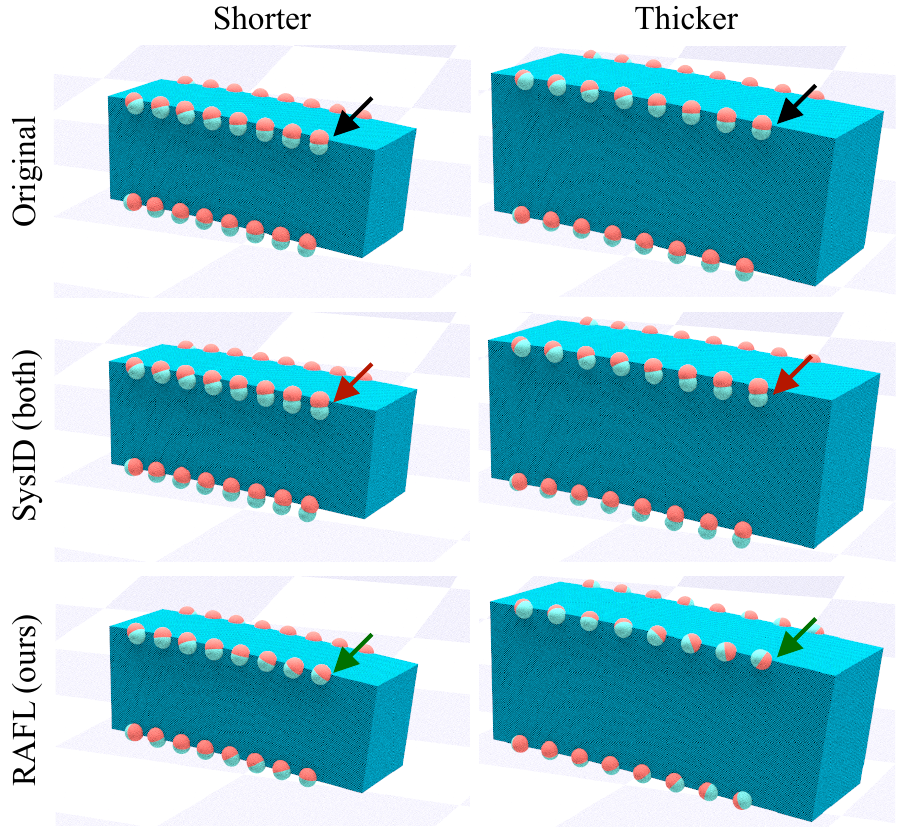} }}
  \caption{Final frames of representative generalization trajectories for non-scaled shorter and thicker geometries. Black arrows indicate the equilibrium configuration produced by the untuned simulator. Red arrows highlight negative transfer in SysID, where the equilibrium configuration deviates further from the real trajectory than the untuned simulation. Green arrows indicate successful zero-shot transfer with RAFL.}
  \label{negative_transfer}
\end{figure}

Fig.~\ref{negative_transfer} compares the final equilibrium configurations for representative trajectories of the shorter and thicker beams. The top row shows the real trajectory, the middle row shows the SysID simulation, and the bottom row shows the RAFL simulation. Black arrows indicate the equilibrium positions predicted by the untuned simulator.
Red arrows highlight the exaggerated stiffness produced by SysID, resulting in beam configurations that deviate further from the real trajectory.
Green arrows indicate that RAFL more closely matches the real equilibrium configuration and reduces marker error when applied zero-shot to unseen geometries.

\end{document}